\documentclass[journal]{IEEEtran}
\usepackage[T1]{fontenc}
\usepackage{graphicx}
\usepackage{times}
\usepackage{helvet}
\usepackage{courier}
\usepackage{amsmath}
\usepackage{algorithm}
\usepackage{algpseudocode}
\usepackage{xcolor}
\usepackage{csquotes} 
\usepackage{color}
\usepackage{paralist}
\usepackage{amssymb}
\usepackage{indentfirst}
\usepackage{subfigure}
\usepackage{float}
\usepackage{multirow}
\usepackage{cite}
\usepackage{mathrsfs}

\usepackage{mathrsfs} 
\usepackage{amsfonts}
\usepackage{todonotes}
\usepackage{pgfplots} 
\usepackage{epstopdf}
\usepackage{epsfig}
\pgfplotsset{compat=newest}
\usepackage{color}
\definecolor{forestgreen}{RGB}{0,139,69}
\usepackage{xcolor}
\definecolor{citecolor}{HTML}{0071bc}
\usepackage[colorlinks, linkcolor=red, anchorcolor=blue, citecolor=citecolor]{hyperref} 

\usepackage{xcolor}
\definecolor{SeaGreen4}{RGB}{0,205,102} 
\definecolor{SlateBlue}{RGB}{106,90,205} 
\definecolor{DarkRed}{RGB}{178,34,34} 
	
\usepackage[switch]{lineno}

\usepackage{textcomp,booktabs}
\usepackage{amssymb}
\usepackage{pifont}

\usepackage{makecell}

\usepackage{colortbl}
\definecolor{mygray}{gray}{.9}
\definecolor{mypink}{rgb}{.99,.91,.95}
\definecolor{mycyan}{cmyk}{.3,0,0,0}

\begin{document}

\title{Adversarial Semantic and Label Perturbation Attack for Pedestrian Attribute Recognition}

\author{Weizhe Kong, Xiao Wang*, \emph{Member, IEEE}, Ruichong Gao, Chenglong Li*, Yu Zhang, \\ Xing Yang, Yaowei Wang, \emph{Member, IEEE}, Jin Tang 
\thanks{$\bullet$ Weizhe Kong, Chenglong Li are with the School of Artificial Intelligence, Anhui University, Hefei, China. (email: weizhekong99@gmail.com, lcl1314@foxmail.com)} 
\thanks{ $\bullet$ Xiao Wang, Ruichong Gao, Jin Tang are with the School of Computer Science and Technology, Anhui University. (email: \{xiaowang, tangjin\}@ahu.edu.cn)} 
\thanks{$\bullet$ Yu Zhang and Xing Yang are with the National University of Defense Technology, Hefei, China. (email: \{zhangyu24d, yangxing17\}@nudt.edu.cn)} 
\thanks{$\bullet$ Yaowei Wang is with Harbin Institute of Technology, Shenzhen, China; Peng Cheng Laboratory, Shenzhen, China. (email: wangyw@pcl.ac.cn)}  
\thanks{* Corresponding Author: Xiao Wang and Chenglong Li} 
}

\markboth{ IEEE Transactions on ***, 2025 } 
{Shell \MakeLowercase{\textit{et al.}}: Bare Demo of IEEEtran.cls for IEEE Journals}

\maketitle

\begin{abstract}
Pedestrian Attribute Recognition (PAR) is an indispensable task in human-centered research and has made great progress in recent years with the development of deep neural networks. However, the potential vulnerability and anti-interference ability have still not been fully explored. To bridge this gap, this paper proposes the first adversarial attack and defense framework for pedestrian attribute recognition. 
Specifically, we exploit both global- and patch-level attacks on the pedestrian images, based on the pre-trained CLIP-based PAR framework. It first divides the input pedestrian image into non-overlapping patches and embeds them into feature embeddings using a projection layer. Meanwhile, the attribute set is expanded into sentences using prompts and embedded into attribute features using a pre-trained CLIP text encoder. A multi-modal Transformer is adopted to fuse the obtained vision and text tokens, and a feed-forward network is utilized for attribute recognition. 
Based on the aforementioned PAR framework, we adopt the adversarial semantic and label-perturbation to generate the adversarial noise, termed ASL-PAR. We also design a semantic offset defense strategy to suppress the influence of adversarial attacks. 
Extensive experiments conducted on both digital domains (i.e., PETA, PA100K, MSP60K, RAPv2) and physical domains fully validated the effectiveness of our proposed adversarial attack and defense strategies for the pedestrian attribute recognition. The source code of this paper will be released on \url{https://github.com/Event-AHU/OpenPAR}. 
\end{abstract}


\IEEEpeerreviewmaketitle

\section{Introduction}

\IEEEPARstart{P}{edestrian} attribute recognition (PAR)~\cite{wang2022PARsurvey} is a basic and important task for human-centric perception, which targets describing the appearance of a given pedestrian image using a set of attributes like "\textit{gender (male, female), hair style, shoes, carrying things}", etc. In addition, it can also be treated as middle-level semantic features and help other human-related tasks, such as person re-identification~\cite{lin2019improving}, multi-object tracking~\cite{li2024attmot}, and pedestrian retrieval~\cite{huang2024attPersonRetrieval}. With the help of deep learning, pedestrian attribute recognition has also been developed rapidly. More in detail, the Convolutional Neural Networks (CNNs) and Recurrent Neural Networks (RNNs) are used for PAR, e.g., the Multi-Task Convolutional Neural Network (MTCNN)~\cite{2015mtcnn} and Joint Representation Learning (JRL)~\cite{wang2017JRL}. Visual-Textual Binding (VTB)~\cite{cheng2022VTB} applies the Transformer to the identification of pedestrian attributes and mines the semantic cues of given pedestrian attributes.

\begin{figure}
\centering
\includegraphics[width=1\linewidth]{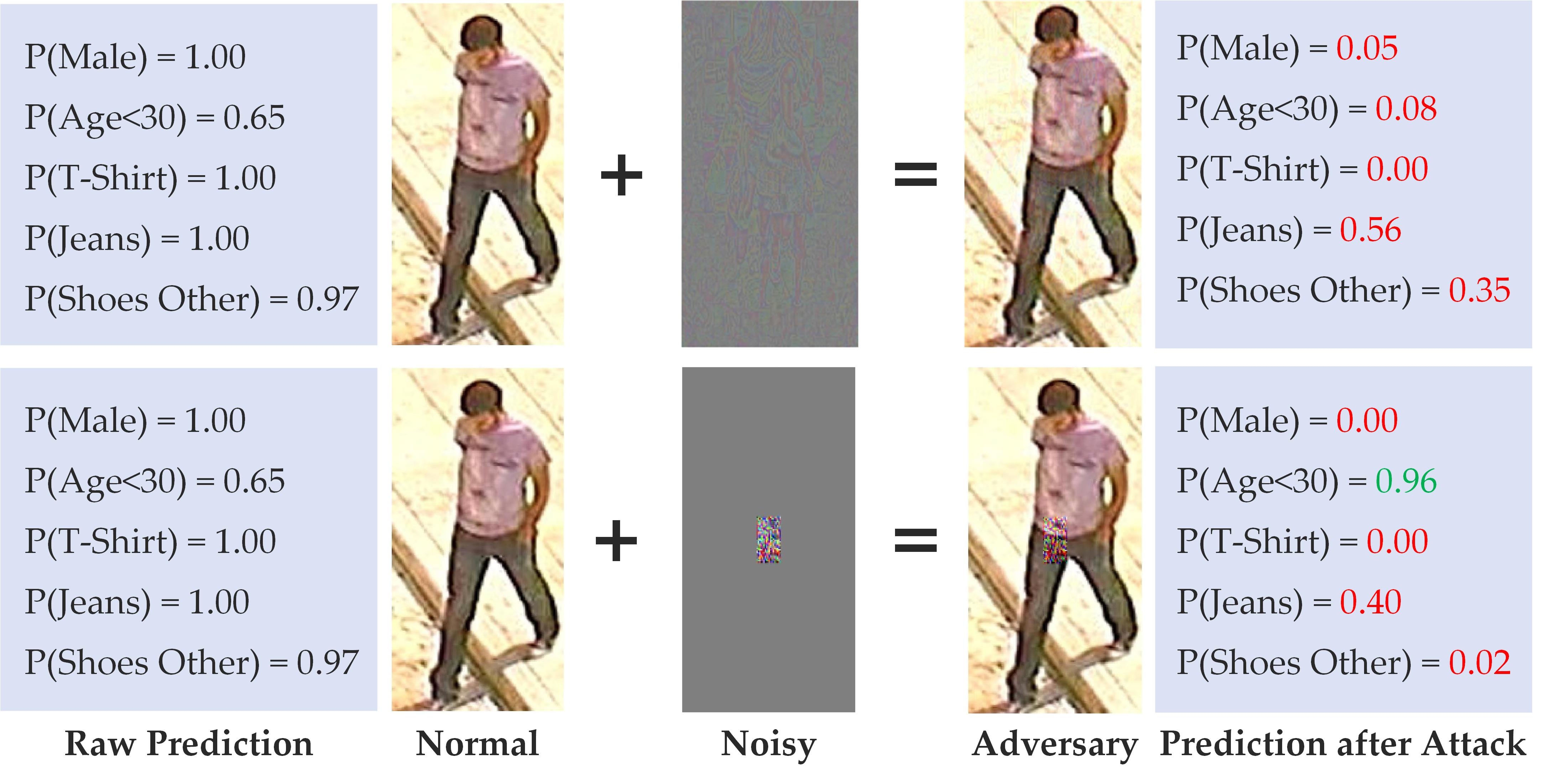}
\caption{Visualization of Global/Local Adversarial Attack for the Pedestrian Attribute Recognition.} 
\label{fig:attackVIS}
\end{figure}

In recent years, the large-scale pre-trained foundation models, such as CLIP~\cite{radford2021CLIP}, GPT-4~\cite{achiam2023gpt}, DeepSeek~\cite{guo2025deepseek}, Qwen~\cite{bai2023qwen}, have drawn more and more attention due to their success in natural language processing, multi-modal understanding. Some researchers also exploit these foundation models for the pedestrian attribute recognition task, for example, Wang et al. propose the PromptPAR~\cite{wang2023PromptPAR} by prompting the pre-trained CLIP model for the PAR task. Jin et al. propose the LLM-PAR~\cite{jin2024pedestrian}, which is the first work to introduce a large language model to augment the attribute recognition. Despite the progress of pedestrian attribute recognition, however, the security issues of PAR have received little attention, especially the adversarial attacks and defense problems~\cite{wei2024physicalAttackSurvey} targeting deep neural networks. As shown in Fig.~\ref{fig:attackVIS}, the performance of a strong PAR model~\cite{wang2023PromptPAR} can be dropped significantly under the attack of global-/local-perturbations.

\begin{figure*}
\centering
\includegraphics[width=1\linewidth]{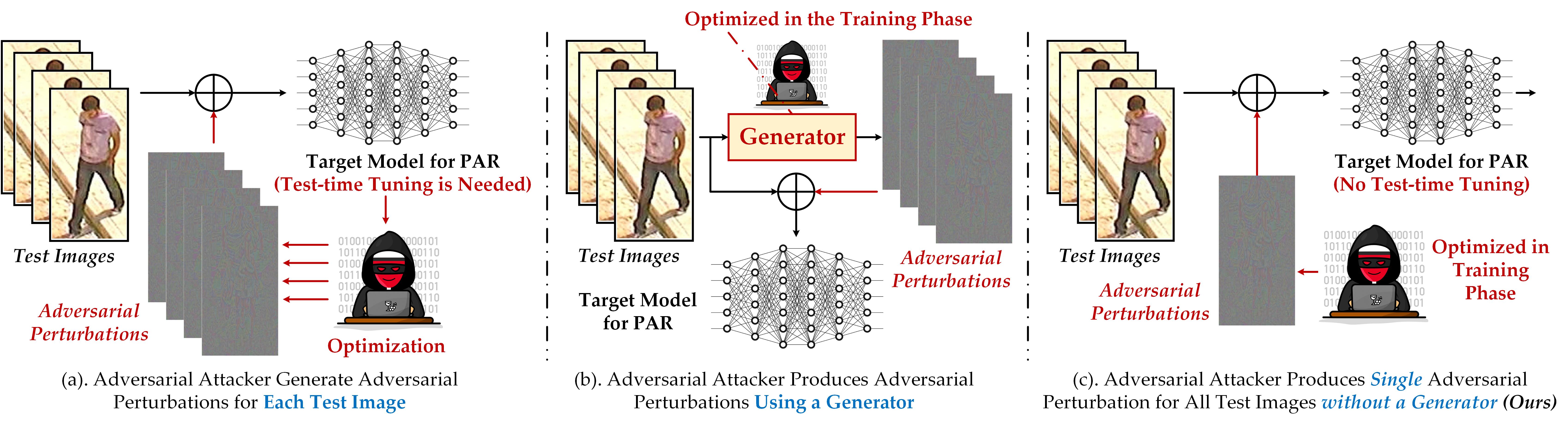}
\caption{Comparison between existing adversarial attackers (a, b) and our newly proposed one (c).} 
\label{fig:demonstration}
\end{figure*}

Actually, adversarial attack and defense have been widely exploited in the deep learning community. Specifically, 
Ye et al.~\cite{ye2024mutual} propose an adversarial attack method based on generator patterns and employ an iterative optimization strategy to attack or defend visual input in text input. 
Kang et al.~\cite{kang2023diffattack} use a backward diffusion process to significantly enhance the semantic plausibility and transfer across models of perturbations by exploiting the generative mechanism of score matching. Fang et al.~\cite{fang2024strong} propose a random adversarial attack process instead of the original deterministic attack process initialized from an input through random image transformation, effectively improving the generalization of adversarial samples. Directly employing existing adversarial attack strategies to fool the PAR models can be a simple attempt, however, these works may be limited by the following issues: 
1). The gradient-based instance-level methods, as shown in Fig.~\ref{fig:demonstration} (a), require multiple gradient calculations for each image during the testing process to iteratively eliminate noise and obtain adversarial samples. This method is time-consuming and lacks generalizability. 
2). The generator-based methods need to train a generator with a relatively large number of parameters during the training stage, as shown in Fig.~\ref{fig:demonstration} (b), which needs more computational resources. 
Therefore, it is urgent and necessary to exploit efficient and effective adversarial attack and defense strategies for the pedestrian attribute recognition task.

In this paper, we study the adversarial attack strategy based on PromptPAR~\cite{wang2023PromptPAR} for pedestrian attribute recognition. Given the pedestrian image and attribute phrase, the PromptPAR encodes them using the CLIP visual and textual encoder, respectively. Then, a multi-modal Transformer network is adopted to fuse the two features. A feed-forward network is adopted to transform the multi-modal features into attribute predictions. To build a universal and generator-free adversarial attack strategy, in this work, we propose an adversarial semantic and label perturbation based on the gradient cues, termed ASL-PAR. For the \textit{label perturbation}, its basic working principle is to optimize an adversarial noise by calculating the loss between the model's predictions and the perturbed labels, and then adding this noise to the input pedestrian image to fool the PAR model into outputting incorrect results. The design of \textit{semantic perturbation} mainly focuses on the misalignment of visual-textual features, thereby enhancing the attack effectiveness of adversarial noise. In order to ensure the invisibility of the attack, we select the $L_{\infty}$ norm from~\cite{liu2024stealthiness} as the constraint  index, and set the $L_{\infty}$ norm of the noise to be no more than 10/255. In addition, we also develop a defense strategy to suppress the influence of disturbances. An overview of our proposed adversarial attack and defense framework can be found in Fig.~\ref{fig:framework} and Fig.~\ref{fig:defensefranmework}.

To sum up, the contributions of this paper can be summarized as the following three aspects: 

$\bullet$ We exploit the adversarial attack strategy for the pedestrian attribute recognition, which pioneers the study of security issues in PAR models, laying a solid foundation for future research. 

$\bullet$ We propose simple but effective semantic and label-perturbation adversarial attack strategies that significantly decrease overall PAR performance, termed ASL-PAR. These results highlight the vulnerability of PAR models, and it is worth devoting more research efforts to the security of PAR. 

$\bullet$ To mitigate adversarial attack noise's impact, we propose a corresponding defense strategy based on the PromptPAR model, which fully underscores the importance of the defense strategy.

In addition, extensive experiments in both the digital domain (including PETA~\cite{deng2014peta}, PA100K~\cite{liu2017hydraplus}, RAP-V2~\cite{2019rapv2}, and MSP60K~\cite{jin2024pedestrian} datasets) and physical domain demonstrate the effectiveness of the proposed strategy.

\textbf{\textit{The rest of this paper is organized as follows:}} 
In Section~\ref{sec::relatedWorks}, we give a brief introduction to the related works on pedestrian attribute recognition and adversarial attack \& defense. In Section~\ref{sec::method}, we introduce the adversarial attack and defense algorithm for the PAR. In Section~\ref{sec::experiments}, we conduct the experiments and give the quality and quantity analysis on multiple PAR benchmark datasets. Finally, we conclude this paper and provide the possible research direction in Section~\ref{sec::conclusion}.

\section{Related Works} \label{sec::relatedWorks}

In this section, we review the related works on Adversarial Attack and Defense, as well as Pedestrian Attribute Recognition. More related works can be found in the following survey~\cite{wang2022PARsurvey} and paper list~\footnote{\url{github.com/wangxiao5791509/Pedestrian-Attribute-Recognition-Paper-List}}. 

\subsection{Adversarial Attack and Defense} 
The evolution of adversarial attacks and defenses has been marked by continuous innovation driven by the interplay between increasingly sophisticated attack strategies and corresponding defensive mechanisms. The field originated with the seminal work of Szegedy et al.~\cite{szegedy2013intriguing}, who demonstrated that imperceptible perturbations could mislead deep neural networks (DNNs). This discovery spurred foundational methods like the Fast Gradient Sign Method (FGSM)~\cite{goodfellow2014explaining}, a single-step gradient-based attack, and its iterative variant, I-FGSM~\cite{kurakin2018adversarial}, which refined perturbations through multi-step updates. To address the challenge of black-box transferability, momentum-based techniques such as MI-FGSM~\cite{dong2018boosting} emerged, stabilizing gradient directions through momentum accumulation, while input transformation strategies like the Diverse Input Method (DIM)~\cite{xie2019improving} and Translation-Invariant Method (TIM)~\cite{dong2019evading} enhanced attack generalization through randomized resizing, padding, and gradient approximations over shifted inputs. Parallel advancements in frequency-domain attacks, exemplified by AdvDrop~\cite{duan2021advdrop}, introduced a paradigm shift by crafting adversarial examples through the removal of critical high-frequency details rather than additive perturbations, exposing DNNs' reliance on non-robust features and challenging traditional noise-detection defenses. In addition, Xu et al.~\cite{xu2023best} proposed a minimum pixel attack method to attack the text recognition model, which pointed out that this is a way to protect the text information by attacking the model, which pointed out another application scenario for adversarial attacks.

On the defensive front, adversarial training~\cite{madry2017towards} became a cornerstone, hardening models by training on adversarially perturbed data, while ensemble adversarial training~\cite{tramer2017ensemble} extended robustness by incorporating perturbations from multiple models. Input pre-processing techniques, including JPEG compression~\cite{guo2017countering} and randomization~\cite{xie2017mitigating}, aimed to disrupt adversarial patterns, and feature-level defenses like High-Level Representation Guided Denoising~\cite{liao2018defense} sought to purify inputs in the latent space. In addition, Variance Tuning~\cite{wang2021enhancing} enhanced attack transferability by stabilizing gradients through neighborhood variance analysis, while Composite Transformation Methods~\cite{lin2019nesterov} combined input transformations to bypass advanced defenses. Certified robustness methods like Randomized Smoothing~\cite{cohen2019certified} introduced formal guarantees but faced limitations against adaptive attacks. Uddin et al. ~\cite{uddin2023robust} proposed a defense strategy for generative adversarial attacks, which used multiple generators to generate multiple instances, and used multiple instances to train the model, which effectively restrained the generative attack.

Generative adversarial attacks, which leverage generative models to craft adversarial examples, have evolved significantly since the inception of generative adversarial nets (GANs)~\cite{goodfellow2014generative}. Early approaches focused on integrating GANs with adversarial example generation, such as adversarial perturbation synthesis through generator-discriminator frameworks, enabling scalable and transferable attacks. AdvGAN~\cite{xiao2018generating} pioneered the integration of Generative Adversarial Networks (GANs) to synthesize adversarial perturbations. By training a generator to produce perturbations that both deceive target classifiers and maintain visual similarity, AdvGAN demonstrated real-time adversarial example generation while preserving semantic features. Building on this, AT-GAN~\cite{wang2019gan} further unified adversarial training and attack generation within a GAN-based framework, enabling adaptive adversarial example synthesis across varying defense scenarios. More recently, diffusion models have been explored for generating semantically coherent adversarial samples, addressing limitations in diversity and naturalness. Zhu et al.~\cite{zhu2022label} proposed a method to obtain adversarial samples by reverse-fitting the attacked model only through the prediction results output by the attacked model and taking the prediction results as the input of another model. Meta-GAN~\cite{feng2023robust} proposes a physical adversarial attack method that is not only able to produce robust physical adversarial examples, but also able to adapt to unseen classes and new DNN models by accessing a small number of digital and physical images. Adversarial attacks can pose serious risks to PAR systems, as subtle perturbations can mislead attribute predictions in surveillance and security applications. Studying these vulnerabilities is critical to developing robust PAR models.

\subsection{Pedestrian Attribute Recognition}
Pedestrian Attribute Recognition (PAR) has evolved significantly with advancements in deep learning. 
Early approaches primarily relied on Convolutional Neural Networks (CNNs) for visual feature extraction and multi-label classification~\cite{an2020hierarchical,abdulnabi2015multi}, but their local receptive fields struggled to model long-range dependencies between attributes. 
To address this, Recurrent Neural Networks (RNNs) are introduced to capture contextual correlations through sequential attribute prediction. 
Wang et al.~\cite{wang2017JRL} employed LSTMs to dynamically adjust visual features based on previously predicted attributes, simulating human-like reasoning by leveraging temporal dependencies. 
Zhao et al.~\cite{zhao2019recurrent} extended this idea with convolutional LSTMs, modeling intra-group attribute relationships and inter-group correlations via recursive attention. 
In addition, Graph Neural Networks (GNNs) explicitly encode attribute relationships through graph structures. 
Fan et al.~\cite{fan2020correlation} designed a correlation graph convolutional network (CGCN) that propagates semantic information across co-occurring attributes, addressing data imbalance by sharing features among correlated nodes. 
Tan et al.~\cite{tan2020relation} further introduced relation-aware GCNs to model both spatial dependencies and semantic hierarchies, significantly improving recognition accuracy for rare or ambiguous attributes.

Attention mechanisms further enhanced PAR by focusing on attribute-specific regions, as seen in HydraPlus-Net’s multi-scale feature aggregation~\cite{liu2017hydraplus} and Visual Attention Consistency (VAC) frameworks~\cite{guo2022visual}. 
The advent of Vision Transformers (ViTs) revolutionized PAR by leveraging global self-attention~\cite{dosovitskiy2020image}, exemplified by visual-textual fusion~\cite{cheng2022VTB, Zhu_2023_CVPR, Wang2024SpatioTemporalST} and DRFormer’s dual-relation modeling of spatial and semantic interactions~\cite{tang2022drformer}. 
Wang et al.~\cite{Wang2025RGBEventBP} develop a new RWKV-based multi-modal fusion PAR framework by aggregating the RGB frames and event streams. 
Recent generative paradigms, such as SequencePAR~\cite{jin2023sequencepar}, reformulated PAR as a sequence generation task using masked Transformers to mitigate data imbalance, while PromptPAR~\cite{wang2023PromptPAR} harnessed CLIP’s vision-language pretraining with region-aware prompts for zero-shot generalization. 
Despite progress, the exploration of adversarial attacks in PAR remains nascent compared to their extensive study in generic vision tasks. Traditional PAR models exhibit vulnerability to perturbations targeting localized features. Systematic investigations into PAR-specific adversarial threats and defenses are critical to ensure reliable deployment in safety-critical scenarios.

\begin{figure*}
\centering
\includegraphics[width=1\linewidth]{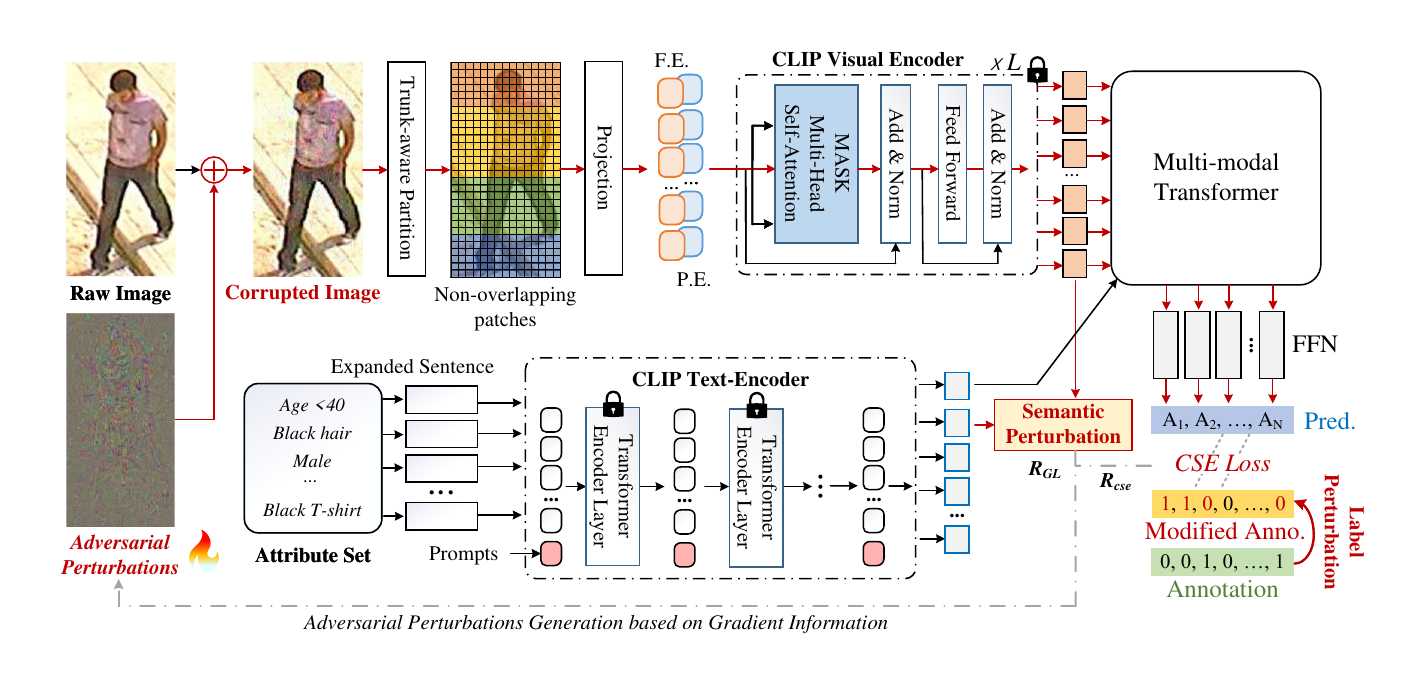}
\caption{An overview of our proposed adversarial attack framework for pedestrian attribute recognition, termed ASL-PAR. Given the pedestrian image and attribute set, the CLIP model is adopted for feature extraction and attribute embedding. The two features are fed into the multi-modal Transformer for attribute recognition. Based on the PAR framework, we design a novel semantic perturbation and label perturbation attack to generate adversarial perturbations. 
} 
\label{fig:framework}
\end{figure*}

\section{Our Proposed Approach} \label{sec::method}

In Section~\ref{baseline}, we first present the baseline we attacked, PromptPAR~\cite{wang2023PromptPAR} proposed by Wang et al. 
In Section~\ref{overview}, we introduce the overview of the proposed adversarial attack and defense method. 
In Section~\ref{attack} and Section~\ref {defense}, we dive into the details of these two strategies, respectively, and introduce the loss functions used in this paper in Section~\ref{loss}.

\subsection{Preliminary: PromptPAR} \label{baseline} 
PromptPAR~\cite{wang2023PromptPAR} is a CLIP-based pedestrian attribute recognition (PAR) framework that formulates the PAR as a multi-modal fusion problem. It consists of four main modules, i.e., a vision encoder, a text encoder, a Multi-modal Transformer (MM-Former), and a classification head. By encoding the input pedestrian image and the attributes with the help of pre-trained CLIP~\cite{radford2021CLIP}, PromptPAR well captures the vision-language relations. Specifically, the CLIP uses ViT-L/14 for visual feature encoding, which divides the pedestrian image into equal-sized patches as input. For the attribute phrases, it is first segmented into separate words, and then each attribute is expanded into a sentence using the prompt template. The CLIP text encoder takes sentences as input and then outputs linguistic feature representations. After obtaining the visual and textual features, they are concatenated together and fed into a multi-modal Transformer. Meanwhile, they adjust the prompts to generate the visual region features that match the attribute representations well using the global-local image text similarity aggregator~\cite{abdelfattah2023cdul}. Finally, a classification head is introduced to project the obtained features into the corresponding attribute response scores. Different from the conventional full fine-tuning for neural network optimization, PromptPAR optimizes network parameters as few as possible through prompt tuning to prevent overfitting and preserve the vision-language feature space of CLIP as much as possible.

\subsection{Overview}~\label{overview} 
As shown in Fig.~\ref{fig:framework}, our proposed adversarial attack strategy is developed based on the PromptPAR framework. Specifically, it can be divided into two parts. The first part is to offset the true value of the label, such as the description of the upper body's dress, and adjust the original true value of the attribute to the attribute of another upper body's dress, to perturb the attribute on the pedestrian part. Then, the perturbed attribute labels are used to train the fully frozen PromptPAR model against noise. The second part is to semantically align the image features and text features of CLIP output according to the perturbed labels on the basis of label perturbation, to realize semantic interference in the CLIP feature space. Thus enhancing the effect of the attack. Our attack method is tested on public PAR datasets and achieves a good attack effect. 

In addition, we also design a corresponding defense method against our proposed attack method. The defense approach is also divided into two parts. On the one hand, since noise is added at the input of the image, we correspondingly add a filter in front of the image encoder. The main function of this filter is to recover the pictures containing attack noise to the state not attacked by learning as much as possible, to achieve the effect of filtering noise. However, although this filter has a significant effect on the attack of adding noise to the input, it still cannot restore the model to its original performance. Therefore, given the interference with the semantic space of CLIP in the method, we add the adjustment of the Prompt to the learnable state on the text input side, to align the semantic space of CLIP by fine-tuning the Prompt. From the experimental results, the fine-tuning on the text makes the model's performance further recover, making the model's performance as close as possible to the non-attack state.

\subsection{Pedestrian Part Semantic Offset Adversarial Attack} \label{attack} 
Given a pedestrian image $ \mathcal{I} \in R^{H \times W \times C}$, the goal of pedestrian attribute recognition is to predict a set of attributes the person has from a pre-defined attribute list $\mathcal{A} = \{A_1, A_2, ..., A_N\}$. The $H, W,$ and $C$ are the height, width, and channel of image $\mathcal{I}$, $N$ is the number of all the defined attributes.

\noindent $\bullet$ \textbf{Input Process.~} 
For the PromptPAR model, we directly load its trained weights and then freeze it completely. The main calculation process is as follows: first, in the input part, the pedestrian image $\mathcal{I}$ is added with a randomly initialized trainable adversarial noise $\eta$ of the same size, and the adversarial sample $\mathcal{I}_{Noisy}$ is obtained, i.e., 
\begin{equation}
    \mathcal{I}_{Noisy} = \mathcal{I} + \eta
\end{equation}

The adversarial noise in this paper is shared by all samples of the same dataset. Then, the adversarial samples $\mathcal{I}_{Noisy}$ are fed into the image encoder to obtain the image features $\mathcal{F}_{img}$. The input of text is the attribute set $\mathcal{A}$ which is expanded into sentences $\mathcal{S} = \{S_1, S_2, ..., S_N\}$ through the template and directly fed into the text encoder to obtain text features $\mathcal{F}_{text}$. Therefore, we have:
\begin{align}
&\mathcal{F}_{img} = ImageEncoder(\mathcal{I}_{Noisy}) \label{eq:ie},\\
&\mathcal{F}_{text} = TextEncoder(\mathcal{S})\label{eq:te}
\end{align}

After obtaining the image feature $\mathcal{F}_{img}$ and the text feature $\mathcal{F}_{text}$, the two features are concatenated and sent to the multi-modal Transformer for modal interaction. And get $\mathcal{F}_{fusion}^{img}$ and $\mathcal{F}_{fusion}^{text}$ after the multi-modal Transformer. $\mathcal{F}_{fusion}^{text}$ is taken as the discriminant feature fed into the prediction head to obtain the predicted probability value of a set of attributes $Prob = \{p_1, p_2, ..., p_N\}, p_k\in [0,1]$. Thus, we get:
\begin{align}
&\mathcal{F}_{fusion}^{img},\mathcal{F}_{fusion}^{text} = MM\text{-}Former (\mathcal{F}_{img},\mathcal{F}_{img})\label{eq:mmformer},\\
&Prob = Sigmoid(FFN(\mathcal{F}_{fusion}^{text})) \label{eq:heads}
\end{align}
Finally, we take the predicted values of the set of attributes and calculate the loss, and then apply the gradient inversion to update the adversarial noise we added to the image.

\noindent $\bullet$ \textbf{Label Perturbation.~} 
Our proposed attack method belongs to the target attack, a specific error label is set. We cut in from the perspective of pedestrian parts and perturb the attributes of the same part, as shown in Fig.~\ref{fig:framework}. Firstly, we introduce some symbolic identities, the set of truth labels $\mathcal{L}_{true} = \{L_1, L_2, ..., L_N\}$, where $L_N$ takes the value 0 or 1. These attributes can be roughly divided into gender attribute $\mathcal{L}_g =\{L_1\}$, head attributes $\mathcal{L}_h = \{L_2, ..., L_h\}$, upper body attributes $\mathcal{L}_u = \{L_{h+1}, ..., L_u\}$, lower body attributes $\mathcal{L}_l = \{L_{u+1}, ..., L_l\}$, foot attributes $\mathcal{L}_f = \{L_{l+1}, ..., L_f\}$, and hand attributes$\mathcal{L}_{h'} = \{L_{f+1}, ..., L_N\}$. Here, $h, u, l, f$ represent the boundaries of each set of attributes. It is followed by the implementation of the attack method.

According to the division of pedestrian parts mentioned above, we disturb the internal attributes of each part. This is done by randomly shifting an attribute with value 1 in the truth-valued attribute set, which is present in the image, to another attribute in the same part group that is not present. If the attribute group has only one attribute, such as gender, then it is negated. If the number of real attributes in an attribute group is greater than the number of non-real attributes, more real attributes will be randomly retained. The following example is provided to better understand this procedure. 
\begin{equation}
\begin{split}
\mathcal{L}_{true} = \{\mathcal{L}_g, \mathcal{L}_h, &\mathcal{L}_u, \mathcal{L}_l, \mathcal{L}_f, \mathcal{L}_{h'} \},\\
\mathcal{L}_g = \{0\} &\Rightarrow \{\textcolor{red}{1}\} = \mathcal{L}_g^{p},\\
\mathcal{L}_h = \{0,1,0\} &\Rightarrow \{\textcolor{red}{1,0,\textcolor{black}{0}}\} = \mathcal{L}_h^{p},\\
\mathcal{L}_u = \{0,0,0,1,1\} &\Rightarrow \{\textcolor{red}{1\textcolor{black}{,0,}1,0,0}\} = \mathcal{L}_u^{p},\\
\mathcal{L}_l = \{1,1,1,0,0\} &\Rightarrow \{1,\textcolor{red}{0,0,1,1}\} = \mathcal{L}_l^{p},\\
\mathcal{L}_f = \{0,1,0\} &\Rightarrow \{\textcolor{red}{1,0},0\} = \mathcal{L}_f^{p},\\
\mathcal{L}_{h'} = \{0,0,0\} &\Rightarrow \{0,0,0\} = \mathcal{L}_{h'}^{p},\\
\mathcal{L}_{perturbation} = \{\mathcal{L}_g^p, &\mathcal{L}_h^p, \mathcal{L}_u^p, \mathcal{L}_l^p, \mathcal{L}_f^p, \mathcal{L}_{h'}^p \} 
\end{split}
\end{equation} 
where $\mathcal{L}_{true}$ represents the true attribute label and $\mathcal{L}_{perturbation}$ represents the perturbed label. The $\Rightarrow$ in the middle represents the process of randomly shifting labels. In this way, the pressure of learning against noise can be reduced, and the prediction results after the attack can be in the range of reasonable prediction errors, which is not easily detected.

\begin{algorithm}[t]
\caption{Adversarial Attack on PromptPAR} 
\label{parattackcode}
\small 
\raggedright 
\textbf{Input}: Original RGB Image $\mathcal{I}$, Original Attribute Sentence $\mathcal{S} = \{S_1, S_2, ..., S_N\}$, Perturbation $\eta$\\
\textbf{Parameter}: Total training epoch $E$, perturbation size $\epsilon$, loss weight $\alpha$, model parameters $\theta$ \\
\textbf{Output}: Perturbation $\eta^{E}$ after training for $E$ epochs 
\begin{algorithmic}[1]
 \State \textcolor{blue}{\% Initialize}
 \State Randomly initialize the perturbation $\eta^{1}$
 \For{$e = 1$ to $E$}
 \State \textcolor{blue}{\% Forward Propagation} 
 \State $\mathcal{I}^{e}_{Noisy} \leftarrow$ add $\eta^{e}$ to $\mathcal{I}$
 \State Get $\mathcal{F}_{img}, \mathcal{F}_{text}$ from Equ.~\eqref{eq:ie} and Equ.~\eqref{eq:te}
 \State Get $\mathcal{F}_{fusion}^{img}, \mathcal{F}_{fusion}^{text}$ from Equ.~\eqref{eq:mmformer}
 \State Get $Prob$ from Equ.~\eqref{eq:heads}
 \State \textcolor{blue}{\% Update Noise} 
 \State Compute $R_{cse} = \nabla L_{cse}({\mathcal{I}^{e}_{noisy}}, \theta)$ according to Equ.~\eqref{cse_loss}
 \State Compute $R_{GL} = \nabla L_{GL}({\mathcal{I}^{e}_{noisy}}, \theta)$ according to Equ.~\eqref{gl_loss}
 \State Update $\eta^{e+1} \leftarrow Clip(|\eta^{e}-(R_{cse}+\alpha \cdot R_{GL})|<\epsilon)$ 
 \EndFor
 \State \Return $\eta^{E}$
\end{algorithmic}
\end{algorithm}

\noindent $\bullet$ \textbf{Semantic Perturbation.~} 
In this paper, in addition to perturbing the labels, we also perturb the feature space of CLIP accordingly. The specific approach is to use the perturbed label as the matching truth value, and to narrow the features from the image features $\mathcal{F}^{img}_{GL}$ and the text features $\mathcal{F}^{text}_{GL}$ according to the perturbed label. The features of the text and the image are further extracted by the similarity aggregator~\cite{abdelfattah2023cdul}, as shown in the following equation:
\begin{align}
&\mathcal{F}^{img}_{GL}, \mathcal{F}^{text}_{GL} = Aggregator(\mathcal{F}_{img}, \mathcal{F}_{text}), \\
&P_{GL} = \frac{\mathcal{F}^{img}_{GL} \cdot \mathcal{F}^{text}_{GL}}{||\mathcal{F}^{img}_{GL}|| \cdot ||\mathcal{F}^{text}_{GL}||}
\end{align}
where $||*||$ denotes the L2 norm of the vector, $P_{GL}=\{p_{00}^{GL}, p_{01}^{GL}, ..., p_{MN}^{GL}\}$ represents the similarity score between $F_{GL}^{img}$ and $F_{GL}^{text}$, where M is the total number of samples and N is the total number of attributes. 

In this way, the adversarial noise can start from perturbing the features from the CLIP encoder, to perturbing the image features and text features in the fusion, in a progressive way to learn how it affects the model performance.

\noindent $\bullet$ \textbf{Patch-level Attack.~} 
We also try a patch-level scaling on the scale of noise. This is done by randomly initializing a $3 \times 30 \times 30$ noise block and then adding it to the central region of the original image. The following specific computational optimization is consistent with the global noise. The main reason for trying this way is that global noise is difficult to apply to real attacks in the physical world, while this patch-level noise can be printed and attached to the human body to carry out attacks. The experiments in Sub-section~\ref{sec::advPhysic} also verify the effectiveness of this method in physical domain attacks.

\subsection{Defending Against Semantic Offset} \label{defense} 

\begin{figure}
\centering
\includegraphics[width=1\linewidth]{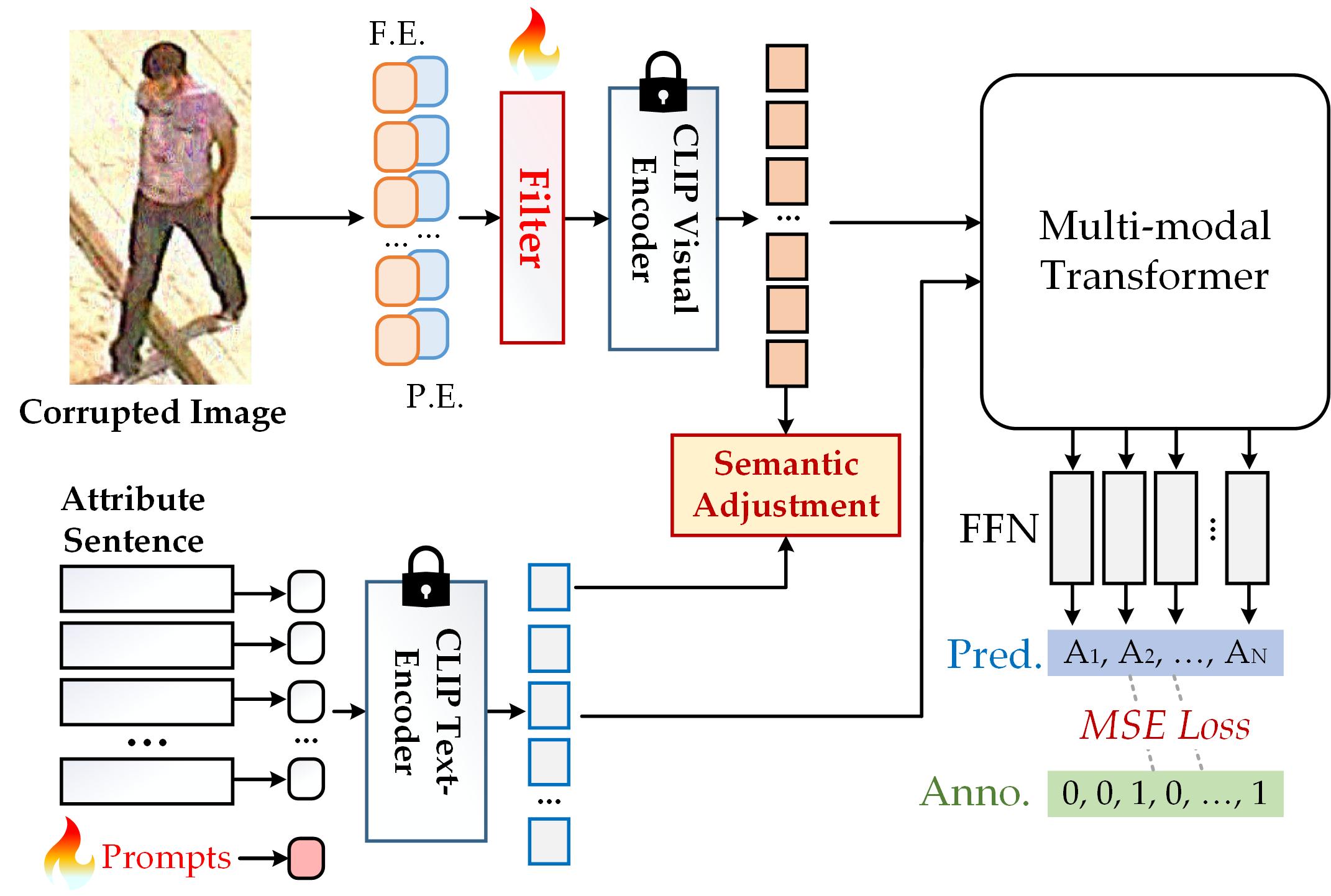}
\caption{An overview of our proposed defense strategy for pedestrian part semantic adversarial attacks.} 
\label{fig:defensefranmework}
\end{figure}

For the proposed adversarial attack on pedestrian parts, we provide a corresponding solution, which is a defense method against the semantic offset of pedestrian parts, as shown in Fig.~\ref{fig:defensefranmework}. First of all, we use a common and effective solution for dealing with adversarial noise added to the input, that is, adding a filter before feeding the input image into the vision encoder. The filter in this paper consists of a convolutional layer, which keeps the dimension of the output image constant relative to the input image. In addition, to further alleviate the interference of adversarial noise on the CLIP semantic space, we add a learnable prompt on the text side. Therefore, the text semantic features can be adjusted to align with the image features again in the CLIP semantic space.

\subsection{Head Network and Loss Function} \label{loss} 
After the effective feature representation is obtained by using the CLIP encoder and the MM-former module, the features are input into the pedestrian attribute recognition prediction head for multi-label prediction. Specifically, our prediction head consists of multiple Feed-Forward Networks (FFNs). The pedestrian attribute features are input into FFNs, and the category corresponding to each input is output to complete the final prediction. As shown in the Equ.~\eqref{eq:heads}, and $Prob=\{p_1, p_2,\dots, p_N \}$, $N$ is the number of attributes. We use the weighted cross-entropy loss as the loss function for our prediction, which can be formulated as: 
\begin{equation}
\label{cse_loss}
\mathcal{L}_{cse}= -\frac{1}{M} \sum_{i=1}^M \sum_{j-1}^N w_j(y_{ij}\log(p_{ij})+(1-y_{ij})\log(1-p_{ij}))
\end{equation}
where $w_j= e^{-r_j} $ is the imbalance weight of the j$^{th}$ attribute, $r_j$ is the proportion of occurrences of the j$^{th}$ attribute in the training subset. $p_{ij}$ and $y_{ij}$ denotes predicted attributes and ground truth, respectively. Then we aim at the interference of the semantic space of CLIP, using the GL loss~\cite{wang2023PromptPAR} proposed by Wang et al., the specific process is as follows: 
\begin{equation}
\label{gl_loss}
 \mathcal{L}_{GL} = -\frac{1}{M} \sum_{i=1}^M \sum_{j-1}^N (y_{ij}\log(p_{ij}^{GL})+(1-y_{ij})\log(1-p_{ij}^{GL}))
\end{equation}
where $p_{ij}^{GL}$ represents the Similarity between image and text features obtained by Similarity Aggregator~\cite{abdelfattah2023cdul}. The final overall loss is computed as follows:
\begin{equation}
 \mathcal{L} =\mathcal{L}_{cse} + \alpha\mathcal{L}_{GL}
\end{equation}
where $\alpha$ is a tradeoff parameter between the two loss functions, and we experimentally set it as 0.5 in our experiments. The overall process can be seen in Algorithm~\ref{parattackcode}.

\begin{table*}[!htb]
\center
\small 
\caption{Comparison with state-of-the-art methods on PETA and PA100K datasets. The first and second are shown in red and blue, respectively. 
} 
\label{petaandpa100kresult}
\begin{tabular}{c|c|c|ccccc|ccccc}
\hline \toprule [0.5 pt] 
\multicolumn{1}{c|}{\multirow{2}{*}{\textbf{\makecell[c]{Methods}}}} 
&\multicolumn{1}{c|}{\multirow{2}{*}{\textbf{\makecell[c]{Publish}}}}&\multicolumn{1}{c|}{\multirow{2}{*}{\textbf{\makecell[c]{$L_{10}$}}}} & \multicolumn{5}{c|}{\textbf{PETA}} & \multicolumn{5}{c}{\textbf{PA100K}} \\ \cline{4-13} 
\multicolumn{1}{c|}{} &
\multicolumn{1}{c|}{} &
 \multicolumn{1}{c|}{} &
 \multicolumn{1}{c}{mA} &
 \multicolumn{1}{c}{Acc} & 
 \multicolumn{1}{c}{Prec} &
 \multicolumn{1}{c}{Recall}&
 \multicolumn{1}{c|}{F1} &
 \multicolumn{1}{c}{mA} &
 \multicolumn{1}{c}{Acc} & 
 \multicolumn{1}{c}{Prec} &
 \multicolumn{1}{c}{Recall}&
 \multicolumn{1}{c}{F1} \\ \hline
 PromptPAR~\cite{wang2023PromptPAR}& TCSVT24 & & 88.76 & 82.84 & 89.04 & 89.74 & 89.18 & 87.47& 83.78 & 89.27& 91.70&90.15 \\ \hline 
 FGSM~\cite{goodfellow2014explaining}& arXiv14 & \checkmark & 63.09 & 48.07 & 61.23 & 64.34 & 62.15 & 59.44 &46.28 &60.05 &62.42 &60.00 \\
 MIFGSM~\cite{dong2018boosting}& CVPR18 & \checkmark & 63.21 & 48.23 & 61.32 & 64.48 & 62.26 &61.08 & 53.97&70.25 &67.10 &67.90 \\
 PGD~\cite{madry2018towards} & ICML18 & \checkmark & 62.37 & 47.27 & 60.56 & 63.40 & 61.37 & 60.52& 53.07& 69.62&66.21 &67.14 \\
 GRA~\cite{zhu2023boosting} & ICCV23 &\checkmark & 49.40 & 29.87& 40.76 & 49.03 & 44.18 &51.90&38.98 &57.54 &52.45 &54.13 \\
 NCS~\cite{qiu2024enhancing}& arXiv24 & \checkmark&\textcolor{blue}{48.20}& 29.38& \textcolor{blue}{40.37}& 48.63 & 43.79 & \textcolor{red}{50.31}&\textcolor{black}{36.68} &\textcolor{blue}{55.25} &\textcolor{black}{50.12} &51.85 \\
 ANDA~\cite{fang2024strong} & CVPR24& \checkmark& \textcolor{red}{47.59} &\textcolor{red}{24.35} &\textcolor{red}{35.45} & \textcolor{red}{40.67}&\textcolor{red}{37.56} &\textcolor{black}{51.17}&\textcolor{red}{31.91} &\textcolor{red}{49.72} &\textcolor{red}{44.12}&\textcolor{red}{46.07} \\
ASL-PAR (Ours) &- & \checkmark & \textcolor{black}{48.25} & \textcolor{blue}{29.02} & 46.72 & \textcolor{blue}{41.54} & \textcolor{blue}{43.29} & \textcolor{blue}{50.85}& \textcolor{blue}{36.34} & 56.10 & \textcolor{blue}{49.37} &\textcolor{blue}{51.42} \\
\hline \toprule [0.5 pt] 
\end{tabular} 
\end{table*}

\section{Experiments} \label{sec::experiments}

\subsection{Datasets and Evaluation Metric} 
In our experiments, the PromptPAR model is tested against attacks mainly on the PETA~\cite{deng2014peta}, PA100K~\cite{liu2017hydraplus}, RAP-V2~\cite{2019rapv2}, and MSP60K~\cite{jin2024pedestrian} datasets. In addition, we design a local adversarial noise attack to obtain a small range of adversarial noise using our attack. We then printed these patterns and conducted physical domain attacks. Let's briefly describe these datasets one by one: 

\noindent $\bullet$ \textbf{PETA Dataset}~\cite{deng2014peta}~\footnote{\url{https://mmlab.ie.cuhk.edu.hk/projects/PETA.html}} contains 19,000 outdoor or indoor pedestrian images and 61 binary attributes. These images are divided into 9500 as the training subset, 1900 as the validation subset, and 7600 as the test subset. In the experiment, we selected 35 pedestrian attributes as prediction targets according to the method of~\cite{deng2014peta}.

\noindent $\bullet$ \textbf{PA100K Dataset}~\cite{liu2017hydraplus}~\footnote{\url{https://github.com/xh-liu/HydraPlus-Net}} is the largest pedestrian attribute recognition dataset, which contains 100,000 pedestrian images and 26 binary attributes. In our experiments, we split them into a training and validation set of 90,000 images and a test subset of the remaining 10,000 images.

\noindent $\bullet$ \textbf{MSP60K Dataset}~\cite{jin2024pedestrian}~\footnote{\url{https://github.com/Event-AHU/OpenPAR/tree/main/MSP60K_Benchmark_Dataset}} is a newly released pedestrian attribute recognition dataset, which contains 60122 pedestrian images and 57 binary attributes. In the experiment, according to the setting of MSP60K, 30,298 images are used as the training set, 6002 images are used as the validation set, and 23,822 images are used as the test set in the random partition. In the cross-domain partition, 34128 images from five scenes (construction site, market, kitchen, school, ski resort) are used for training, and 24994 images from three scenes (Outdoors1, Outdoors2, Outdoors3) are used for testing.

\noindent $\bullet$ \textbf{RAP-V2 Dataset}~\cite{2019rapv2}~\footnote{\url{https://www.rapdataset.com/rapv2.html}} has 84,928 pedestrian images and 69 binary attributes, of which 67,943 are used for training. We selected 54 attributes to train and evaluate our model. 

\noindent $\bullet$ \textbf{Real-world Dataset}~\footnote{\url{https://github.com/Event-AHU/OpenPAR/tree/main/AttackPAR}} is taken by three different people, respectively, in the situation of no holding against noise and holding against noise. To better test the attack performance, we held the hand in front, behind, and to the side in three directions for shooting. Correspondingly, our normal photo is also a picture of these three viewpoints. A total of 139 images are taken and labeled, including 12 normal images. The remaining 127 are images holding adversarial noise.

To verify the effectiveness of our proposed adversarial attack model, we adopt three evaluation metrics widely used in pedestrian attribute recognition tasks to measure the attack effect, including \textbf{mA}, \textbf{Accuracy}, \textbf{Precision}, \textbf{Recall}, and \textbf{F1-measure}. Specifically, the mA can be expressed as: 
\begin{equation} 
\textit{mA} = \frac{1}{N}\sum_{i=1}^{N} AP_{i}
\end{equation}
where $AP_{i}$ is the area under the precision-recall curve for the $attribute_{i}$, and $N$ is the total number of attributes. 
The Accuracy can be expressed as:
\begin{equation}
\textit{Accuracy} = \tfrac{TP\:+\:TN}{TP\:+\:TN\:+\:FP +\:FN}
\end{equation}
where TP is a positive sample predicted correctly, TN is a negative sample predicted correctly, FP is a negative sample predicted incorrectly, and FN is a positive sample predicted incorrectly. 
The Precision, Recall, and F1-measure can be expressed as: 
\begin{align}
&Precision = \tfrac{TP}{TP\:+\:FP}, \\
&Recall = \tfrac{TP}{TP\:+\:FN}, \\
&F1 = \tfrac{2 \times Precision \times Recall}{Precision\:+\:Recall} 
\end{align}

\begin{table*}[!htb]
\center
\small 
\caption{Comparison with state-of-the-art methods on MSP60K and RAPv2 datasets. The first and second are shown in red and blue, respectively.} 
\label{mspandrapresult}
\begin{tabular}{l|c|c|ccccc|ccccc}
\hline \toprule [0.5 pt] 
\multicolumn{1}{c|}{\multirow{2}{*}{\textbf{\makecell[c]{Methods}}}} 
&\multicolumn{1}{c|}{\multirow{2}{*}{\textbf{\makecell[c]{Publish}}}}&\multicolumn{1}{c|}{\multirow{2}{*}{\textbf{\makecell[c]{$L_{10}$}}}} & \multicolumn{5}{c|}{\textbf{MSP60K}} & \multicolumn{5}{c}{\textbf{RAPv2}} \\ \cline{4-13} 
\multicolumn{1}{c|}{} &
\multicolumn{1}{c|}{} &
 \multicolumn{1}{c|}{} &
 \multicolumn{1}{c}{mA} &
 \multicolumn{1}{c}{Acc} & 
 \multicolumn{1}{c}{Prec} &
 \multicolumn{1}{c}{Recall}&
 \multicolumn{1}{c|}{F1} &
 \multicolumn{1}{c}{mA} &
 \multicolumn{1}{c}{Acc} & 
 \multicolumn{1}{c}{Prec} & 
 \multicolumn{1}{c}{Recall}&
 \multicolumn{1}{c}{F1} \\ \hline 
 PromptPAR~\cite{wang2023PromptPAR}& TCSVT24 & & 63.24&53.62&66.15&71.84&68.32 & 83.14 & 69.62 & 77.42 & 85.73 & 81.00 \\ \hline
 FGSM~\cite{goodfellow2014explaining}& arXiv14 & \checkmark & 52.39 & 37.49 &51.16&57.20 &53.32 & 55.87 & 42.90 & 67.97 & 52.52 & 57.75 \\
 MIFGSM~\cite{dong2018boosting}& CVPR18 & \checkmark &52.41 &37.65 & 51.59 &56.19 & 53.33 & 55.85&42.86 &67.94 &52.49 &57.72 \\
 PGD~\cite{madry2018towards} & ICML18 & \checkmark & 52.56 & 38.01 & 52.34 & 56.20 & 53.58 & 55.42 &42.07 & 67.22 &51.72 &56.97 \\
 GRA~\cite{zhu2023boosting} & ICCV23 &\checkmark & 47.12 & 23.93 & 35.95 & 41.34& 37.68 & \textcolor{blue}{50.04}& 30.76& 53.51& 41.63&45.64 \\
 NCS~\cite{qiu2024enhancing}& arXiv24 & \checkmark& \textcolor{red}{46.98}& \textcolor{blue}{24.35} & \textcolor{blue}{37.08} & \textcolor{blue}{40.95} & \textcolor{blue}{38.23} &\textcolor{red}{49.17}&\textcolor{blue}{28.61} &\textcolor{blue}{50.78} & \textcolor{red}{39.64} &\textcolor{blue}{43.27} \\
 ANDA~\cite{fang2024strong} & CVPR24& \checkmark& \textcolor{blue}{47.04} & \textcolor{red}{22.56} & \textcolor{red}{33.47} & \textcolor{red}{40.72}& \textcolor{red}{35.82} & 51.46 & \textcolor{red}{27.29} & \textcolor{red}{43.44}&\textcolor{blue}{41.41} & \textcolor{red}{41.78}\\
ASL-PAR (Ours) &- & \checkmark & 50.78 & 33.20 & 51.09 & 47.49 & 48.58 & 51.71 &37.37 &55.38 &53.02 &53.04 \\
\hline \toprule [0.5 pt] 
\end{tabular} 
\end{table*}

\subsection{Implementation Details} 
The input pedestrian images fed into the CLIP visual encoder are resized to $3 \times 224 \times 224$. During training, we apply random cropping and horizontal flipping for data augmentation. Our implementation uses the ViT-L/14 architecture from CLIP~\cite{radford2021CLIP} as the vision encoder, paired with a text encoder comprising 12 Transformer layers with 512-dimensional hidden states and 8 attention heads. The system employs 50 visual cue tokens and 3 textual cue tokens. We optimize the model for 40 epochs using SGD~\cite{sutskever2013importance} with an initial learning rate of 8e-3. We set the warmup phase to 5 epochs based on the cosine learning rate scheduler. The initial learning rate decreases with a ratio of 0.01 during the warmup phase, and the weight decays to le-4. Training is conducted with a batch size of 32. Additional implementation details are available in our source code.

\subsection{Comparison on Public Benchmark Datasets} 

In this sub-section, we focus on introducing the experimental results on the four widely used PAR benchmark datasets, including PETA, PA100K, MSP60K, and RAPv2. When training the noise, we added the $L_{10}$ constraint to ensure the concealment of the adversarial samples. $L_{10}$ means that the $L_{\infty}$ norm of the noise is at most 10/255.

\noindent $\bullet$ \textbf{Results on PETA Dataset.} 
As shown in Table~\ref{petaandpa100kresult}, our method achieves competitive results on the PETA dataset, matching or even surpassing the latest methods, such as ANDA~\cite{fang2024strong}, NCS~\cite{qiu2024enhancing} in terms of mA and F1 metrics. Moreover, our method does not require gradient backpropagation on the test set and employs a universal adversarial perturbation for all input images, enabling black-box attack capabilities. On the PETA dataset, the performance of the target model, PromptPAR~\cite{wang2023PromptPAR}, for mA, Accuracy, Precision, Recall, and F1 metrics are 88.76, 82.84, 89.04, 89.74, and 89.18, respectively. Our proposed attack reduces PromptPAR's performance on these metrics to 48.25(\textcolor{red}{$\downarrow 40.51$}), 29.02(\textcolor{red}{$\downarrow 53.82$}), 46.72(\textcolor{red}{$\downarrow 42.32$}), 41.54(\textcolor{red}{$\downarrow 48.20$}), 43.29(\textcolor{red}{$\downarrow 45.89$}), where the data is also presented in the order of mA, Accuracy, Precision, Recall and F1 metrics. The efficacy of our proposed method is comparable to that of the latest approaches, demonstrating its excellent attack capability.

\noindent $\bullet$ \textbf{Results on PA100K Dataset.} 
As shown in the Table~\ref{petaandpa100kresult}, our proposed method also demonstrates competitive results on the PA100K dataset. The performance of the target model, PromptPAR, on this dataset for mA, Accuracy, Precision, Recall, and F1 metrics are 87.47, 83.78, 89.27, 91.70, and 90.15, respectively. Our proposed attack reduces PromptPAR's performance on these metrics to 50.85 (\textcolor{red}{$\downarrow 36.62$}) and 36.34 (\textcolor{red}{$\downarrow 47.44$}), 56.10 (\textcolor{red}{$\downarrow 33.17$}), 49.37 (\textcolor{red}{$\downarrow 42.33$}), 51.42(\textcolor{red}{$\downarrow 38.73$}), where the data is also presented in the order mA, Accuracy, Precision, Recall and F1 metrics. These results indicate that our method is competitive with NCS~\cite{qiu2024enhancing}. However, while a performance gap remains when compared to ANDA~\cite{fang2024strong}, our method's efficacy is considered acceptable for cross-dataset black-box attack scenarios. Future work could explore incorporating diverse techniques to further enhance the attack's effectiveness.

\begin{table*}[!htb]
\center
\small 
\caption{Ablation Study of the key modules of our proposed adversarial attack strategies. 
} 
\label{attckablation}
\begin{tabular}{ccccc|ccccc}
\hline \toprule [0.5 pt] 
\multicolumn{1}{c}{\multirow{2}{*}{\textbf{\makecell[c]{Patch \\ Level}}}} &
\multicolumn{1}{c}{\multirow{2}{*}{\textbf{\makecell[c]{Global \\ Level}}}} &
\multicolumn{1}{c}{\multirow{2}{*}{\textbf{\makecell[c]{$L_{10}$}}}} & \multicolumn{1}{c}{\multirow{2}{*}{\textbf{\makecell[c]{Semantic \\ Perturbation}}}} &\multicolumn{1}{c|}{\multirow{2}{*}{\textbf{\makecell[c]{Label \\ Perturbation}}}} & \multicolumn{5}{c}{\textbf{PETA}} \\ 
\cline{6-10} 
\multicolumn{1}{c}{} &
\multicolumn{1}{c}{} &
\multicolumn{1}{c}{} &
\multicolumn{1}{c}{} &
 \multicolumn{1}{c|}{} &
 \multicolumn{1}{c}{mA} &
 \multicolumn{1}{c}{Acc} & 
 \multicolumn{1}{c}{Prec} &
 \multicolumn{1}{c}{Recall}&
 \multicolumn{1}{c}{F1} \\ \hline
 & & & & & 88.76 & 82.84 & 89.04 & 89.74 & 89.18 \\
 \checkmark& & & \checkmark & \checkmark & 52.16 & 34.16 & 50.61 & 48.99 & 49.03 \\ \hline
&\checkmark& & & \checkmark & 47.27 & 28.20 & 50.94 & 37.90 & 42.62 \\
&\checkmark& & \checkmark & \checkmark & 46.51 & 28.64 & 49.29 & 39.51 & 43.18 \\
&\checkmark& \checkmark & & \checkmark & 49.05 & 30.63 & 47.83 & 44.11 & 45.20 \\
&\checkmark& \checkmark & \checkmark & \checkmark & 48.25 & 29.02 & 46.72 & 41.54 & 43.29 \\
\hline \toprule [0.5 pt] 
\end{tabular} 
\end{table*}

\noindent $\bullet$ \textbf{Results on MSP60K Dataset.} 
To evaluate our method's effectiveness in cross-domain scenarios, this paper utilizes a cross-domain data partition. As shown in Table~\ref{mspandrapresult}, the performance of the target model, PromptPAR, on the MSP60K dataset for mA, Accuracy, Precision, Recall, and F1 metrics is 63.24, 53.62, 66.15, 71.84, and 68.32, respectively. Our proposed attack reduces PromptPAR's performance on these metrics to 50.78 (\textcolor{red}{$\downarrow 12.46$}) and 33.20 (\textcolor{red}{$\downarrow 20.42$}), 51.09 (\textcolor{red}{$\downarrow 15.06$}), 47.49 (\textcolor{red}{$\downarrow 24.35$}), 48.58(\textcolor{red}{$\downarrow 19.74$}), where the data is also presented in the order mA, Acc, Prec., Recall and F1 metrics. These experimental results demonstrate that our attack method maintains a degree of effectiveness even in cross-domain settings. However, a significant performance gap exists when compared to state-of-the-art methods. The main reason is that these attack methods are directly based on the test set data, so they do not have cross-domain issues. Such approaches, however, typically require access to the model's gradients during testing, characteristic of a white-box attack setting. In contrast, our method operates as a black-box attack at test time, which constitutes a key advantage.

\noindent $\bullet$ \textbf{Results on RAPv2 Dataset.} 
As shown in the Table. \ref{mspandrapresult}, our proposed method also yields competitive results on the RAPv2 dataset. The performance of the target model, PromptPAR, on this dataset for mA, Accuracy, Precision, Recall, and F1 metrics are 83.14, 69.42, 77.42, 85.73, and 81.00, respectively. Our proposed attack reduces PromptPAR's performance on these metrics to 51.71 (\textcolor{red}{$\downarrow 31.43$}) and 37.37 (\textcolor{red}{$\downarrow 32.25$}), 55.38 (\textcolor{red}{$\downarrow 22.04$}), 53.02(\textcolor{red}{$\downarrow 32.71$}), 53.04(\textcolor{red}{$\downarrow 27.96$}), where the data is also presented in the order mA, Acc, Prec., Recall and F1 metrics. The results show the performance gap between our method and state-of-the-art approaches is larger on RAPv2 than on PA100K. This is somewhat counterintuitive, as the RAPv2 dataset's richer attribute set is theoretically expected to be more conducive to perturbing pedestrian part labels. However, we further analyze the attribute set of RAPv2 and find that it contains the attributes of more abstract concepts, such as the occupation of pedestrians. These abstract attributes can semantically overlap with more localized part-based attributes. This overlap may cause unintended interference during our semantic perturbation process, as modifications targeting one attribute might inadvertently affect others, thereby limiting the attack's overall performance. Consequently, datasets similar to PETA, where attributes are predominantly localized to specific pedestrian parts, are expected to be more amenable to our attack strategy and thus yield better results.

\subsection{Global Noise Visualization in Digital Domain} 
\begin{figure}
\centering
\includegraphics[width=0.49\textwidth]{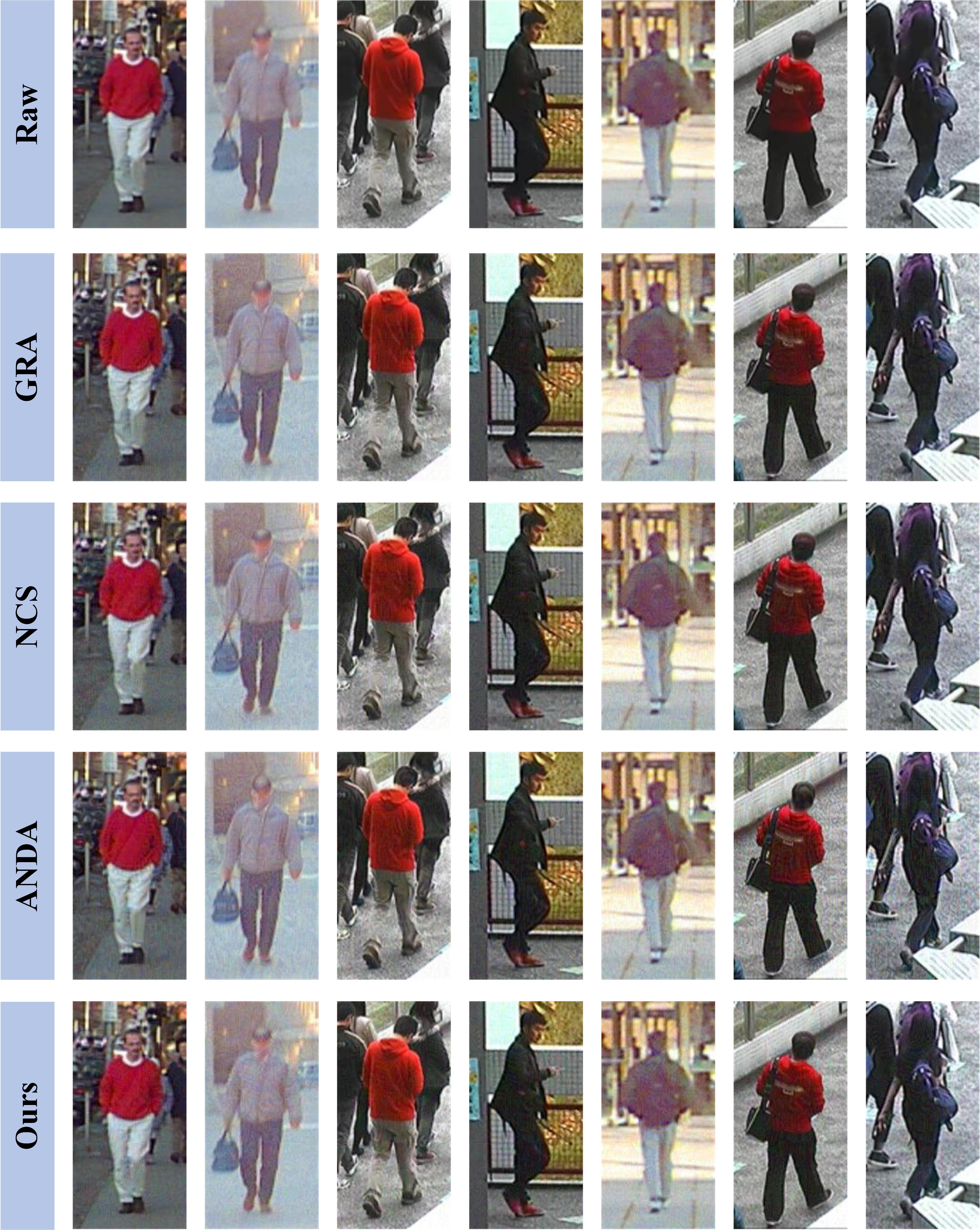}
\caption{Global noise visualization results of the proposed attack method and other attack methods on the PETA dataset.} 
\label{fig:globalnoise}
\end{figure}

In this paper, we visualize the adversarial examples generated by our method and some excellent existing methods on the PETA dataset. The noises used are all generated under the $l_{10}$ constraint, and the visualization results are shown in Fig.~\ref{fig:globalnoise}. It can be seen from the figure that under the constraint of $l_{10}$, neither our method nor the existing method is much different from the original figure, but a careful comparison can find that there are still some differences. However, if there is no original image comparison, it is difficult to find the problem by only observing the adversarial samples. This is also in line with the unobservability of adversarial noise, which makes adversarial attacks not easy to find.

\begin{table*}
\center
\small 
\caption{Demonstrate the effectiveness of cross-dataset attacks. Blue indicates better attack performance.
} 
\label{crossdataset}
\begin{tabular}{cc|ccccc|ccccc}
\hline \toprule [0.5 pt] 
\multicolumn{1}{c}{\multirow{2}{*}{\textbf{\makecell[c]{Cross-Dataset}}}} & \multicolumn{1}{c|}{\multirow{2}{*}{\textbf{\makecell[c]{$L_{10}$}}}} & \multicolumn{5}{c|}{\textbf{PETA->PA100K}}& \multicolumn{5}{c}{\textbf{PETA->MSP60K}}\\ \cline{3-12} 
 \multicolumn{1}{c}{} &
 \multicolumn{1}{c|}{} &
 \multicolumn{1}{c}{mA} &
 \multicolumn{1}{c}{Acc} & 
 \multicolumn{1}{c}{Prec} &
 \multicolumn{1}{c}{Recall}&
 \multicolumn{1}{c|}{F1} &
 \multicolumn{1}{c}{mA} &
 \multicolumn{1}{c}{Acc} & 
 \multicolumn{1}{c}{Prec} &
 \multicolumn{1}{c}{Recall}&
 \multicolumn{1}{c}{F1}\\ \hline 
 &  \checkmark & \textcolor{blue}{50.85} & 36.34 & 56.10 & 49.37 &51.42  & \textcolor{blue}{50.78}& \textcolor{blue}{33.20} & 51.09 &\textcolor{blue}{47.49} &\textcolor{blue}{48.58}\\
  \checkmark &  \checkmark & 52.10 &\textcolor{blue}{29.30} &\textcolor{blue}{40.73} &\textcolor{blue}{47.44} & \textcolor{blue}{43.29}&51.37 & 33.71& \textcolor{blue}{49.06}&50.70 &49.27\\ \hline

\multicolumn{1}{c}{\multirow{2}{*}{\textbf{\makecell[c]{Cross-Dataset}}}} & \multicolumn{1}{c|}{\multirow{2}{*}{\textbf{\makecell[c]{$L_{10}$}}}} & \multicolumn{5}{c|}{\textbf{PETA->RAPv2}}& \multicolumn{5}{c}{\textbf{MSP60K->PETA}}\\ \cline{3-12} 
 \multicolumn{1}{c}{} &
 \multicolumn{1}{c|}{} &
 \multicolumn{1}{c}{mA} &
 \multicolumn{1}{c}{Acc} & 
 \multicolumn{1}{c}{Prec} &
 \multicolumn{1}{c}{Recall}&
 \multicolumn{1}{c|}{F1} &
 \multicolumn{1}{c}{mA} &
 \multicolumn{1}{c}{Acc} & 
 \multicolumn{1}{c}{Prec} &
 \multicolumn{1}{c}{Recall}&
 \multicolumn{1}{c}{F1}\\ \hline 
 &  \checkmark &51.71 & 37.37& 55.38& 53.02& 53.04 &\textcolor{blue}{48.25} &\textcolor{blue}{29.02} &\textcolor{blue}{46.72} &\textcolor{blue}{41.54} &\textcolor{blue}{43.29}\\
  \checkmark &  \checkmark &\textcolor{blue}{51.34} & \textcolor{blue}{31.77}& \textcolor{blue}{48.57}& \textcolor{blue}{47.08}& \textcolor{blue}{46.94}&64.65 &47.08 &57.34 &68.69 &62.08\\ \hline

  \multicolumn{1}{c}{\multirow{2}{*}{\textbf{\makecell[c]{Cross-Dataset}}}} & \multicolumn{1}{c|}{\multirow{2}{*}{\textbf{\makecell[c]{$L_{10}$}}}} & \multicolumn{5}{c|}{\textbf{MSP60K->PA100K}}& \multicolumn{5}{c}{\textbf{MSP60K->RAPv2}}\\ \cline{3-12} 
 \multicolumn{1}{c}{} &
 \multicolumn{1}{c|}{} &
 \multicolumn{1}{c}{mA} &
 \multicolumn{1}{c}{Acc} & 
 \multicolumn{1}{c}{Prec} &
 \multicolumn{1}{c}{Recall}&
 \multicolumn{1}{c|}{F1} &
 \multicolumn{1}{c}{mA} &
 \multicolumn{1}{c}{Acc} & 
 \multicolumn{1}{c}{Prec} &
 \multicolumn{1}{c}{Recall}&
 \multicolumn{1}{c}{F1}\\ \hline 
 &  \checkmark & \textcolor{blue}{50.85}& \textcolor{blue}{36.34}& \textcolor{blue}{56.10}& \textcolor{blue}{49.37}& \textcolor{blue}{51.42} & \textcolor{blue}{51.71}& 37.37 &\textcolor{blue}{55.38} &53.02 &53.04\\
  \checkmark &  \checkmark &55.65 &45.82 &65.88 &57.28 &60.58 &53.61 &\textcolor{blue}{35.29} &64.62 &\textcolor{blue}{42.36} &\textcolor{blue}{49.98}\\ \hline
  
\hline \toprule [0.5 pt] 
\end{tabular} 
\end{table*}

\subsection{Ablation Study}
In this section, the main modules of the attack method and the defense method are analyzed and experimented in detail. The attack methods include the comparison of global and local anti-noise, whether to make $L_{10}$ limit on the influence range of noise, and the effectiveness of pedestrian part semantic interference and pedestrian part label interference. The defense methods include the effectiveness verification of prompt fine-tuning against noise filters and text semantics.

\begin{table}[!htp]
\center
\small 
\caption{Component effectiveness experiments in defense strategies. 
} 
\label{defenseablation}
\resizebox{0.475\textwidth}{!}{
\begin{tabular}{cc|ccccc}
\hline \toprule [0.5 pt] 
\multicolumn{1}{c}{\multirow{2}{*}{\textbf{\makecell[c]{Perturbation \\ Filter}}}} &\multicolumn{1}{c|}{\multirow{2}{*}{\textbf{\makecell[c]{Text Prompt \\ Adjust}}}} & \multicolumn{5}{c}{\textbf{PETA}} \\ \cline{3-7} 
\multicolumn{1}{c}{} &
 \multicolumn{1}{c|}{} &
 \multicolumn{1}{c}{mA} &
 \multicolumn{1}{c}{Acc} & 
 \multicolumn{1}{c}{Prec} &
 \multicolumn{1}{c}{Recall}&
 \multicolumn{1}{c}{F1} \\ \hline
 & & 88.76 & 82.84 & 89.04 & 89.74 & 89.18 \\
 \checkmark & & 85.54 & 77.71 & 85.02 & 87.18 & 85.78 \\
 \checkmark & \checkmark & 85.82 & 78.23 & 85.27 & 87.55 & 86.11 \\
\hline \toprule [0.5 pt] 
\end{tabular} 
}
\end{table}

\noindent $\bullet$ \textbf{Analysis of Attack Method Components.~} 
As shown in Table \ref{attckablation}, firstly, we make a comparison on the size of noise. It can be seen that the effect of noise attack at the Patch level is indeed worse than that of noise attack at the Global level, but this local noise is more conducive to application to the real world. In addition, we also analyze the impact of adding the $L_{10}$ constraint to the noise. According to the experimental results, adding the $L_{10}$ constraint reduces the attack effect slightly. Finally, after further ablation experiments on the two modules in the attack method, it is found that the two modules are effective for the attack method, and the performance under the constraint of $L_{10}$ is very close to the unconstrained state.

\begin{figure*}[!htp]
\centering
\includegraphics[width=1\linewidth]{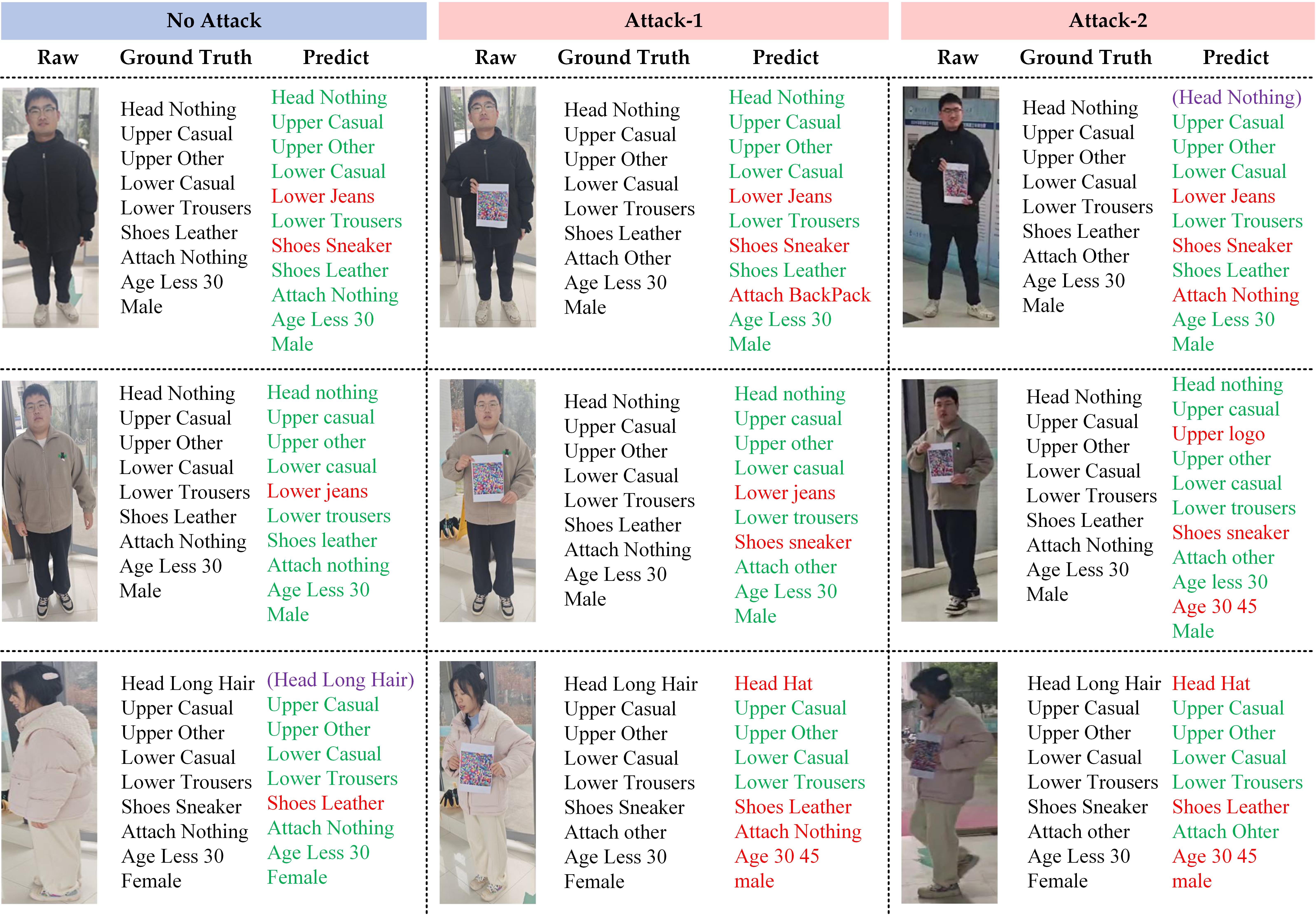}
\caption{Visualization of local noise attacks in the physical domain.} 
\label{fig:physicalworld}
\end{figure*}

\noindent $\bullet$ \textbf{Analysis of Defense Method Components.~} 
As shown in Table \ref{defenseablation}, our proposed targeted defense method demonstrates excellent performance, primarily attributable to the noise filter. This is also because our proposed attack method mainly adds our adversarial attack noise to the original graph. If this noise can be completely filtered out, then comparable performance to the original model can be achieved. However, as the results indicate, employing the filter alone is insufficient to fully restore the model's original performance. This is because the filter can only remove most of the noise, but there will still be a small amount of noise that is difficult to remove. Therefore, the text Prompt fine-tuning strategy proposed by us can further offset the remaining noise influence, so that the model can further recover the performance.

\subsection{Cross-Dataset Attacks} 
We verify the cross-dataset attack capability of our proposed method by conducting cross-dataset attack experiments with noise trained on PETA dataset and noise trained on the MSP60K dataset. As shown in Table~\ref{crossdataset}, the noise trained on PETA data shows excellent results in the other three datasets. In particular, PA100K and RAPv2 go beyond the noise obtained from the dataset's own training. It even surpasses the optimal method ANDA~\cite{fang2024strong} on PA100K dataset. In the PA100K dataset, the mA, Acc, Prec., Recall, and F1 metrics of ANDA are 51.17, 31.91, 49.72, 44.12, and 46.07, respectively. However, the effects of PETA noise cross-dataset attack are 52.10, 29.30, 40.73, 47.44, 43.29. Among them, all the metrics except mA are better than ANDA. This proves that our proposed method also has a good effect on cross-dataset black-box attacks and has the ability of cross-dataset attacks.

Subsequently, we used the noise obtained from the training of MSP60K dataset to conduct cross-dataset attack tests on PETA, PA100K, and RAPv2 datasets, respectively. It is noteworthy that the MSP60K dataset itself exhibits a domain gap: its training set comprises indoor scenes, while its test set features outdoor scenes. These three target datasets (PETA, PA100K, RAPv2) do not have such explicit indoor/outdoor divisions. Specifically, PA100K consists entirely of outdoor scenes, and PETA includes both indoor and outdoor environments. Therefore, applying the MSP60K-trained noise (derived primarily from indoor training data) to these two datasets also constitutes a cross-domain attack scenario. It can also be seen from the experimental results that the effect of MSP60K noise on PETA and PA100K datasets is not ideal, but it is not completely invalid. For the RAPv2 dataset, its data is entirely composed of indoor scenes, so there is relatively no cross-domain problem. So the effect on RAPv2 is better than the other two datasets, and even exceeds the noise of the RAPv2 dataset's training.

In summary, the cross-dataset attack experiments are carried out through the noise of PETA, which verifies that our method has the ability of cross-dataset attacks. In addition, the noise of MSP60K is used to carry out cross-dataset attack experiments, which further verifies the impact of data cross-domain on our proposed method, and provides a basis for subsequent cross-domain attacks.

\subsection{ Adversarial Attack in the Physical Domain } \label{sec::advPhysic}
We made a simple physical world attack performance test dataset. Since the global noise cannot adapt to the physical world attack, we choose the local adversarial noise for testing. But in fact, we also understand that the method of applying global noise to the physical world needs to adjust the shooting equipment. Under the balance, we choose a simpler way to print the local noise. 

The experimental results are shown in Table~\ref{physicalworld}. It can be seen that due to the pure difference between the data we took and the training data, the model's performance is poor. However, the validity of noise can still be found from the experimental results. Moreover, from the visualization results of Fig.~\ref{fig:physicalworld}, it can be found that the noise does have the ability to perturb the attributes in the same part. The purple color represents unpredicted attributes, and the red color represents incorrectly predicted attributes.

\begin{table}
\center
\small 
\caption{Experiments on the effect of adversarial Attacks against Patch-level Noise in the Physical World.} 
\label{physicalworld}
\begin{tabular}{c|ccccc}
\hline \toprule [0.5 pt] 
\multicolumn{1}{c|}{\multirow{2}{*}{\textbf{\makecell[c]{Attack \\ or not}}}} & \multicolumn{5}{c}{\textbf{Real-world dataset}} \\ \cline{2-6}
 \multicolumn{1}{c|}{} &
 \multicolumn{1}{c}{mA} &
 \multicolumn{1}{c}{Acc} & 
 \multicolumn{1}{c}{Prec} &
 \multicolumn{1}{c}{Recall}&
 \multicolumn{1}{c}{F1} \\ \hline
 & 48.06 & 68.78 & 74.99 & 88.08 & 80.78 \\
 \checkmark & 45.58 & 57.72 & 66.95 & 79.13 & 72.30 \\
\hline \toprule [0.5 pt] 
\end{tabular} 
\end{table}

\begin{table}
\center
\small 
\caption{The Attack Performance of the Generated Adversarial Noise on Other Different PAR models.} 
\label{crossmodel}
\resizebox{0.475\textwidth}{!}{
\begin{tabular}{c|c|ccccc}
\hline \toprule [0.3 pt] 
\multicolumn{1}{c|}{\multirow{2}{*}{\textbf{\makecell[c]{Model}}}} & \multicolumn{1}{c|}{\multirow{2}{*}{\textbf{\makecell[c]{Attack}}}} & \multicolumn{5}{c}{\textbf{MSP60K}}\\ \cline{3-7} 
 \multicolumn{1}{c|}{} &
 \multicolumn{1}{c|}{} &
 \multicolumn{1}{c}{mA} &
 \multicolumn{1}{c}{Acc} & 
 \multicolumn{1}{c}{Prec} &
 \multicolumn{1}{c}{Recall}&
 \multicolumn{1}{c}{F1} \\ \hline 
 PromptPAR~\cite{wang2023PromptPAR} & & 63.24&53.62&66.15&71.84&68.32 \\
 PromptPAR~\cite{wang2023PromptPAR} & \checkmark & 50.78 & 33.20 & 51.09 & 47.49 & 48.58 \\ \hline
 VTB~\cite{cheng2022VTB} & &58.59 & 49.81 & 65.11 & 66.11 & 65.00\\
 VTB~\cite{cheng2022VTB} & \checkmark & 57.25& 48.44 & 62.89 & 66.08 & 63.79 \\ \hline
 MambaPAR (Vim)~\cite{wang2024empiricalmamba} & & 56.75 & 47.34 & 61.92 & 64.98 & 62.80 \\
 MambaPAR (Vim)~\cite{wang2024empiricalmamba} & \checkmark & 56.03 & 47.39 & 61.82 & 65.22 & 62.79 \\ 
\hline \toprule [0.5 pt] 
\end{tabular} 
}
\end{table}

\subsection{Limitation Analysis} 
Although the effectiveness of the proposed method is verified on the pedestrian attribute recognition task, the ability of cross-dataset attacks is also verified. However, the ultimate attack goal, that the added adversarial noise be able to cross datasets and models simultaneously, is not achieved. We also conduct experiments with cross-model attacks. The experimental results are shown in Table \ref{crossmodel}. The noise obtained by our method trained on PromptPAR \cite{wang2023PromptPAR} has a significant attack effect on PromptPAR. On the same Transformer architecture model, VTB~\cite{cheng2022VTB}, the performance is moderately affected. However, there is almost no attack effect on MambaPAR~\cite{wang2024empiricalmamba} based on Mamba architecture.
Secondly, through extensive experiments, we find that our attack method is mainly limited by the design of dataset attributes as well as the richness of training data scenarios. 
In the future, it is necessary to break through these limitations to make the attack method more generalized. 


\section{Conclusion} \label{sec::conclusion}
In this paper, we investigate the potential vulnerabilities and robustness of advanced models in pedestrian attribute recognition (PAR) tasks. To the best of our knowledge, this is the first study to explore adversarial attacks within the context of PAR, as well as the first to implement a cross-dataset black-box attack method in this domain. We introduce a label-shifting strategy based on partitioned attribute groups corresponding to different pedestrian body parts, thereby misleading the model through semantic perturbations. These perturbations are further projected into the semantic space of the CLIP visual-textual encoder to compromise model performance. Experimental results demonstrate the effectiveness of both proposed attack strategies. In response to these attacks, we propose a corresponding defense mechanism, which is shown to be highly effective. However, this defense approach relies heavily on a substantial amount of noisy data for training, which poses significant challenges for deployment in real-world scenarios. Thus, developing a defense method that can be efficiently applied in practice remains an important direction for future research. Furthermore, extensive experiments reveal several limitations of the proposed approach, including issues related to attribute set design and domain shifts across datasets. These findings lay a solid foundation for future studies in this area.


\small{ 
\bibliographystyle{IEEEtran}
\bibliography{reference}
}

\end{document}